\documentclass[conference]{IEEEtran}
\usepackage{blindtext, graphicx}
\usepackage{array}
\usepackage{subfigure}
\usepackage{booktabs}
\usepackage{multirow}
\usepackage{amsmath}
\usepackage{array}

\ifCLASSINFOpdf
 
\else

\fi

\begin{document}

\title{Convolutional Neural Networks for Aerial Vehicle Detection and Recognition}
%\titleheader{National Aerospace & Electronics Conference (NAECON) 2018}


\author{\IEEEauthorblockN{Amir Soleimani\IEEEauthorrefmark{1},
Nasser M. Nasrabadi\IEEEauthorrefmark{1}, Elias Griffith\IEEEauthorrefmark{2}, Jason Ralph\IEEEauthorrefmark{2}, Simon Maskell\IEEEauthorrefmark{2}}
\IEEEauthorblockA{\IEEEauthorrefmark{1}Lane Department of Computer Science and Electrical Engineering, West Virginia University\\ \IEEEauthorrefmark{2}Department of Electrical Engineering and Electronics, University of Liverpool \\
Email: a.soleimani.b@gmail.com,
nasser.nasrabadi@mail.wvu.edu\\
e.griffith@liverpool.ac.uk,
jfralph@liverpool.ac.uk,
s.maskell@liverpool.ac.uk}\\
This paper has been accepted in the National Aerospace &\\ Electronics Conference (NAECON) 2018 and would be indexed in IEEE}

\maketitle

\begin{abstract}
This paper investigates the problem of aerial vehicle recognition using a text-guided deep convolutional neural network classifier. The network receives an aerial image and a desired class, and makes a yes or no output by matching the image and the textual description of the desired class. We train and test our model on a synthetic aerial dataset and our desired classes consist of the combination of the class types and colors of the vehicles. This strategy helps when considering more classes in testing than in training.
\\\\
\end{abstract}

\IEEEpeerreviewmaketitle

\section{Introduction}

Aerial imagery, captured by drones or Unmanned Aerial Vehicles (UAVs), is a great tool for surveillance because of its wide field of view and the ability of drones to access places that would otherwise be difficult to visit. Aerial imagery has also other applications like border security, search and rescue tasks, and image and video understanding. Also, it can be used in human-human, human-vehicle, and vehicle-vehicle interaction understanding.

The wide field of view advantages of aerial imagery, however, result in objects of interest occupying small number of pixels in each image. In comparison to the regular view or street view images, aerial images have less information and details about vehicles as well as other objects in the image. Therefore, it is common that a vehicle in an aerial view is missed because of the background or other objects. On the other hand, false positive predictions are also highly probable.

The other issue that makes the resolution challenge harder is the limitation in the computational resources. Although it is possible to take a high resolution image, processing a large image will result in a huge unavoidable computational costs, specially if we are interested in implementing an online aerial vehicle detection system.
\begin{figure}[!t]
\centering
\includegraphics[scale=.85]{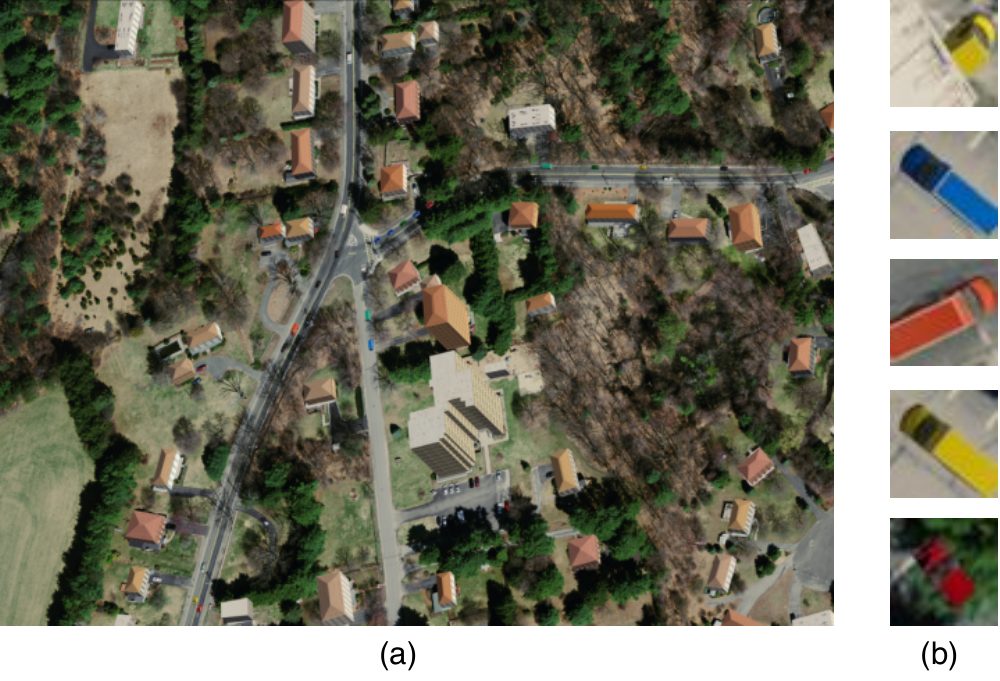}\\
\caption{(a) An aerial image. (b) Some exemplar objects of interest (vehicles).}
\label{fig_graph2}
\end{figure}

The application of aerial vehicle detection and recognition can be more specific if the goal of the system is not just limited to detect vehicles but to find specific vehicles. For example, a detection system can concentrate on searching for a specific car with a specific color, type, and other descriptions (e.g., yellow taxi, large green truck). In this scenario, the detection system can be used in the applications like finding a suspicious vehicle or target vehicle among several other vehicles, objects, and backgrounds.

In this specific goal, in addition to the resolution challenge, there is an open-ended challenge. Providing a comprehensive dataset that covers all the probable objects variations and classes (e.g., vehicle types, colors, shapes, and other variations) is impossible. Therefore, the detection system has to respond to unseen targets. In other words, during the testing phase it sees sample variations that it has not seen in its training phase. This challenge can be called as unseen challenge or open-ended challenge.

Visual Question Answering (VQA) systems take an image and an open-ended textual question about the given image, and try to provide an answer to the question in a textual format \cite{vqapaper}. The core idea behind the VQA task is to answer to unseen questions that could be considered here as classes. Antol et al. by using LSTM and MLP structure achieves acceptable results on their large dataset consisting about 0.25M images, 0.76M questions, and 10M answers\footnote{http://www.visualqa.org/}.

In order to alleviate the challenge of objects occupying small number of pixels, we split the problem into two sub-problems \cite{fusion}. We first assume that a deep detector like Single Shot Multibox Detector (SSD) \cite{ssd} extracts objects or areas of interest, and second, we use a deep convolutional network to recognize which of the extracted objects of interest are also the vehicles we wish to detect.

In this paper, we propose a framework that can handle the problem of open-ended classification or prediction. In a classical image classification system an image is processed and an output label is produced. However, in this paper, we use another novel architecture in which it receives an image and a desired textual description of the class, represented by a code vector, and makes a yes or no decision about the correctness of the input label. In other words, it decides if the input image has the desired textual description of the class label or not (see Figure 2).

\begin{figure}[!t]
\centering
\includegraphics[scale=1.1]{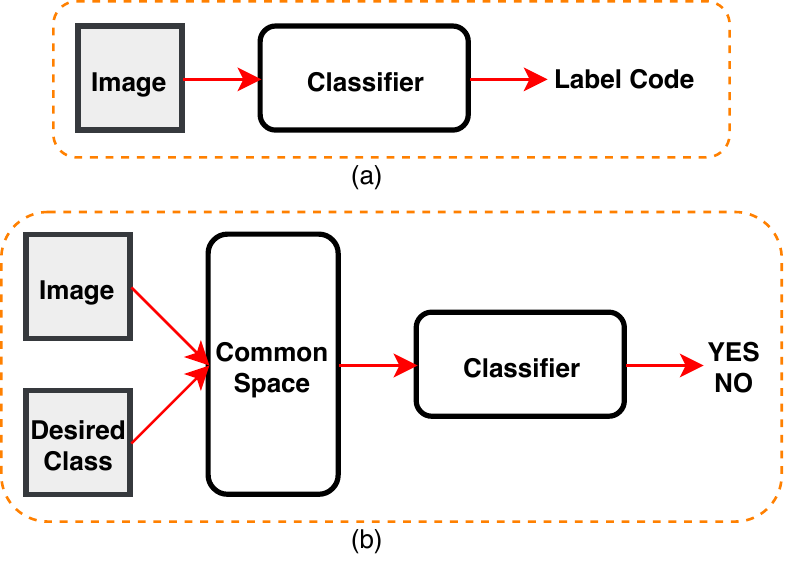}\\
\caption{(a) A classical classifier that receives an image and predicts a label code for the image class. (b) The proposed architecture that can consider classes that have not been seen during the training.}
\label{fig_graph1}
\end{figure}

This paper is organized as follows. In Section II, we review the literature of deep object detectors and visual question answering. Then in Section III, we investigate our proposed framework. In Section IV, we explain the dataset, experiments, and results. Finally, conclusion and future works are explained in the Section V.

\section{Related Works}

The combination of choosing a good hand-engineered image feature descriptors like histogram of gradients (HOG) \cite{hog} and scale-invariant feature transform (SIFT) \cite{sift}, and choosing a classifier like support vector machine (SVM) \cite{svm} or multilayer perceptron (MLP) have been the main focus of research papers in the area of image classification for several years. However, in recent years deep convolutional neural networks, which have outperformed other methods on different datasets like VOC \cite{voc} and Imagenet \cite{imagenet}, are attracting interest in the field of image classification. Deep convolutional neural networks like LeNet \cite{lenet} and AlexNet \cite{alex} have confirmed their effectiveness as classifiers that only receive raw images without any type of image feature descriptors.

R-CNN \cite{RCNN} can be considered as the first considerable and well-known deep structure for the application of object detection. This method takes an input image, then a classical regions of interest generator like selective search \cite{selective} creates about 2000 regions proposals, and a deep convolutional neural network (CNN) extracts visual features, and finally an SVM classifier determines if there is any specific object in these proposals or not. However, doing all these steps separately makes R-CNN slow and inaccurate. They also achieved a good detection result since a high-capacity convolutional neural networks was applied to bottom-up region proposals, and because they used a pre-trained model for their initial points.

Fast R-CNN \cite{fastrcnn} is the next top method in the object detection literature. In contrast to R-CNN, fast R-CNN has its own classifier layer. It takes an image and multiple of regions of interest proposals, then a CNN creates the feature maps for the proposals and a softmax layer and a regressor layer to find the objects in the image. The other important advantage for Fast R-CNN is that it does not need a disk storage for feature caching. 

The next step, which led to real-time object detection with region proposal networks, was Faster R-CNN \cite{faster}. Faster R-CNN eliminates the need for extra regions of interest proposal generator. In other words, generating proposals is also done using a convolutional framework. Their method employs the anchor concept that helps to better catch different objects with different sizes and aspect ratios. It must be noted that Faster R-CNN employs Fast R-CNN for some parts of its algorithm. 

YOLO \cite{yolo} might be considered as the first work in which the object detection problem is considered as a regression problem, while the previous works focused on classifiers to perform detection. A single CNN predicts bounding boxes as well as the class probabilities from only the input images, without any need for the proposals. This is an end-to-end method for detecting objects in the images. While YOLO is well-known for its extreme speed, its accuracy is not as good as the other top methods.\\

Single Shot Multibox detector or SSD \cite{ssd} is another end-to-end single shot object detector that uses a deep learning architecture which in addition to the speed, it has an outstanding accuracy. SSD is a convolutional network that predicts a large number of bounding boxes and scores, which theoretically helps to detect 8732 objects in just one image. The output of SSD, like the previous method, is passed through a non-maximum suppression operator to reduce the overlapping results. SSD is made of a base network, which has a structure like VGG \cite{vgg}. This base network creates a preliminary representation for the next steps which is done using the extra feature layers. These auxiliary layers decrease in size progressively and are responsible for detecting small to large objects. In other words, smaller objects are detected in the earlier layers while the larger ones are detected in the last layers. SSD uses the idea of default boxes, a concept similar to anchors, that helps the network to learn specific filters for specific scales and aspect ratios. These are responsible for objects at different scales and aspect ratios.

\begin{figure*}[!t]
\centering
\includegraphics[scale=1]{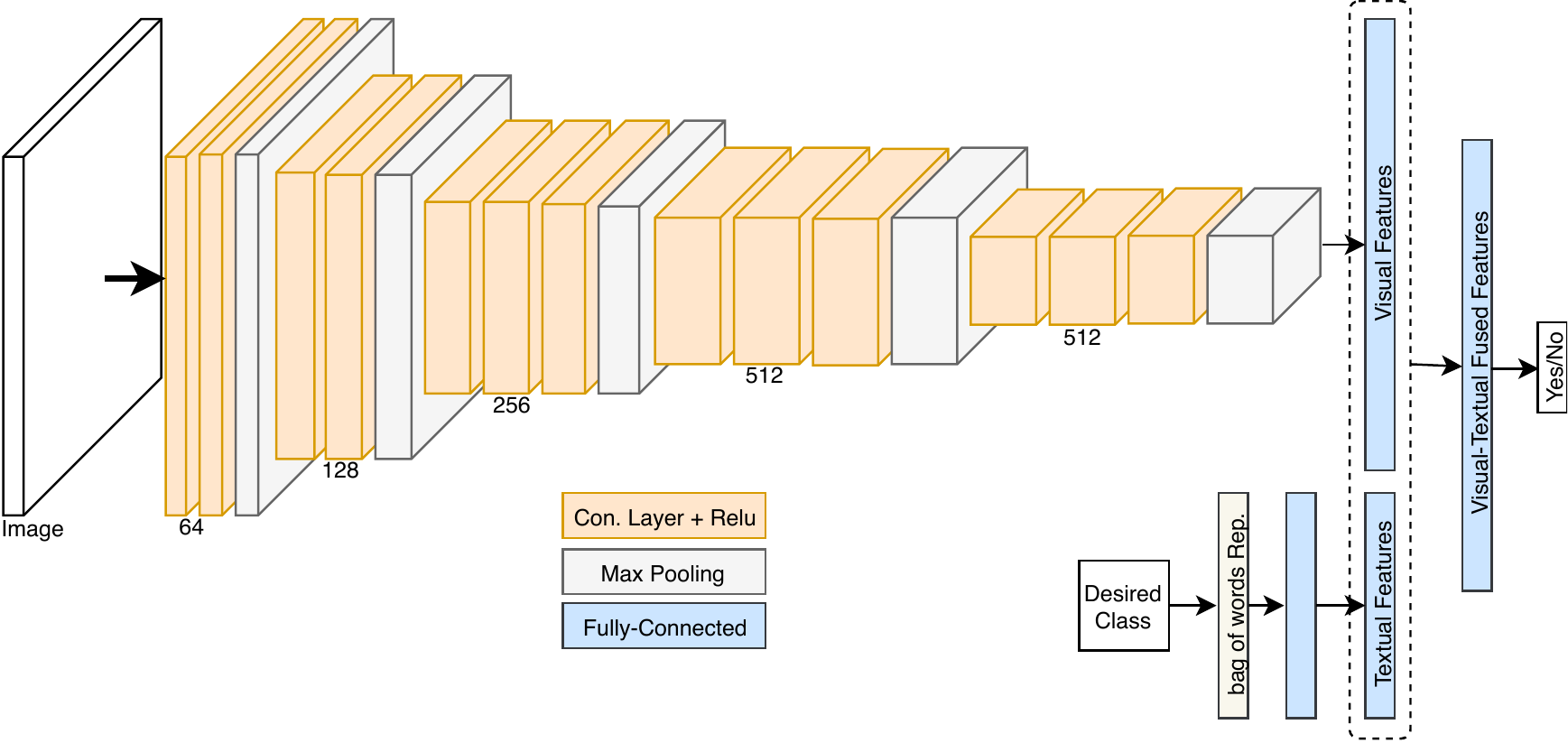}\\
\caption{Proposed method for the task of aerial vehicle detection. Detected objects of interest are described using the features extracted by the convolutional layers. Desired classes are represented using bag of words and then fully connected layers to build a common latent space in which the yes or no decision is made on top of this sub-space.}
\label{fig_graph3}
\end{figure*}

Equation (1) is the SSD's cost function. SSD has an objective function consisting of two parts, one for the localization cost that codes the location of the bounding boxes and the objects, and one for the confidence cost that determines the degree of certainty for the presence of an object in the specific bounding box or location. 

\begin{equation}
l(x,c,l,g)=1/N(L_{conf}(x,c)+{L_{loc}(x,l,g)}).
\end{equation}

The localization cost is a smooth L1 loss and codes the error between the bounding boxes from the ground-truth and the predicted one. In contrast, the confidence cost is a softmax loss that can be considered as a classification cost too. 

Note that $x$ is the parameter that determines the presence of an object in the SSD's default boxes. During the training, this information can be calculated using the ground-truth information and the default boxes positions. $c$ is for the class parameters, and $l$ and $g$ are the parameters for the predicted locations and ground-truth locations, respectively.\\

\section{Proposed Method}

We propose a framework in which, first, an SSD \cite{ssd}, which has shown promising performances in the aerial image object detection literature \cite{fusion} and \cite{okutama}, generates a number of objects of interest proposals for an input aerial image. These proposals might contain vehicle, background, or other objects. In other words, we would have a set of object patches or object chips per original image.

In contrast to the classical structures of the classifiers that receive an input image and predict the labels at the output, we propose an architecture that receives an image as well as the textual description of the desired class as the inputs. This architecture predicts a yes and no decision that shows if the image has the desired class label or not (see Figure 2). One of the main reasons for changing the structure is that in this new structure, the classifier is not limited to pretrained or predefined class labels. In other worlds, class labels are also used like open-ended images.

We use a VGG-16 architecture \cite{vgg} with only one fully connected layer to extract visual descriptors. This convolutional structure consists of five convolutional layers with the following details. The first two layers consist of two similar layers with the depth of 64 and 128, respectively. The next three layers have three similar layers with the depth of 256, 512, and again 512, respectively. Just after each layer, there is a max pooling structure to reduce the spatial size and increase the generalization. A fully connected layer is used just after the fifth layer (see Figure 3) and its values are fed into the next step which is a fully-connected layer for fusion of visual features extracted by the VGG structure and the textual features that is described below.

Meanwhile, the textual description of the desired classes are coded using the bag of words representation, and then a fully connected layer transforms this information into the next space. The textual description of the desired classes in our experiments consist of the color and the types of the vehicles, but they can be more complicated with more details about the vehicles.

As Figure 3 shows, the visual features, which are extracted by the VGG network and the textual features, which are extracted by the fully-connected layers, are fused and form a visual-textual sub-space. This is done by using a fully-connected fusion layer. Finally, a two-class softmax classifier is placed on top of the last layer and trained to predict if the image has the desired class or not (Figure 3 shows the details about the proposed method).

It is worth noting that all the weights of the network, including the visual feature extractor, the textual feature extractor and the softmax classifier are optimized together. This fact will force each component of our algorithm to influence the optimization of other components.\\

\section{Experiments}
In order to evaluate our proposed framework, we use our dataset that contains real aerial images and synthetic cars and trucks which are placed on the streets (see Figure 1 and Figure 4). Vehicles can have seven different colors: black, white, gray, yellow, green, blue, and red. The two types of vehicles in conjunction with these seven colors describe a 14-class recognition problem.

The synthetic aerial dataset contains about 5000 vehicles with information about the vehicle types and colors. See Figure 4 for some examples. More details about the dataset are provided in the following subsection. \\

\begin{figure}[!t]
\centering
\includegraphics[scale=0.3]{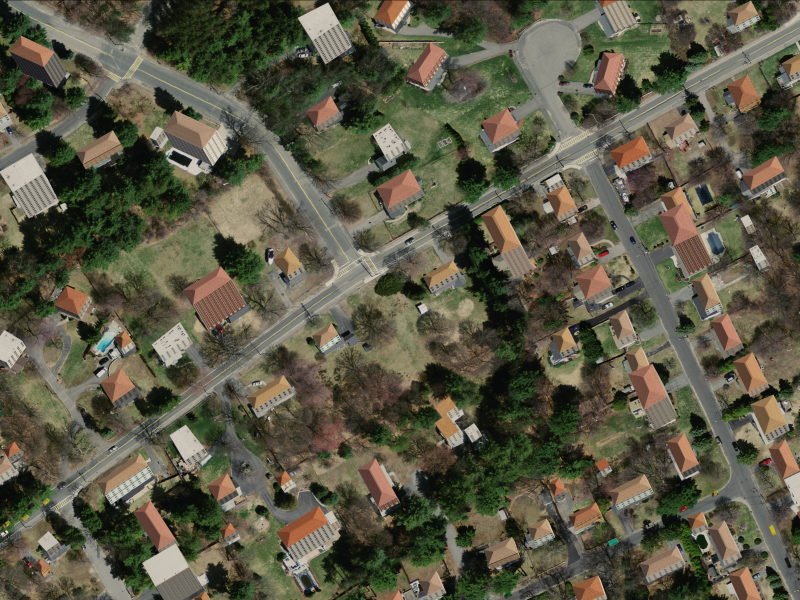}
\\
\vspace{5mm}
 \includegraphics[scale=.3]{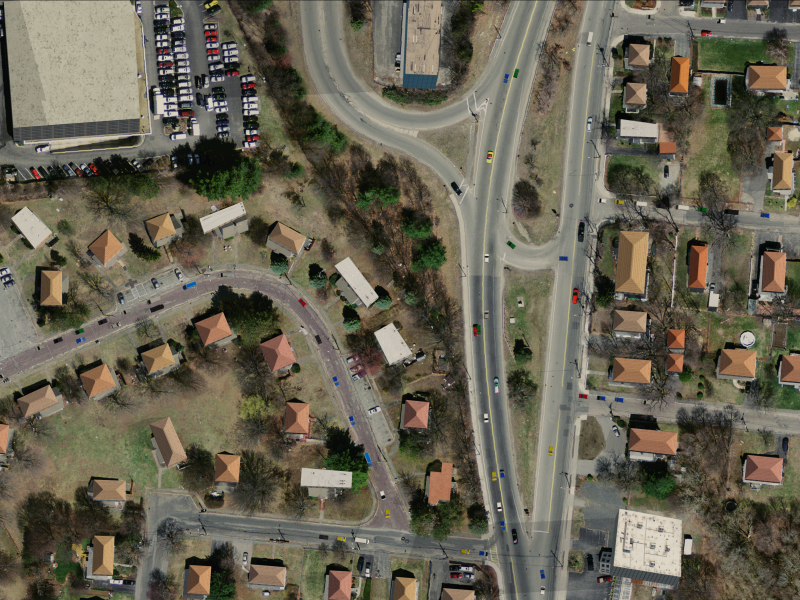}
 \\
\vspace{5mm}
 \includegraphics[scale=.3]{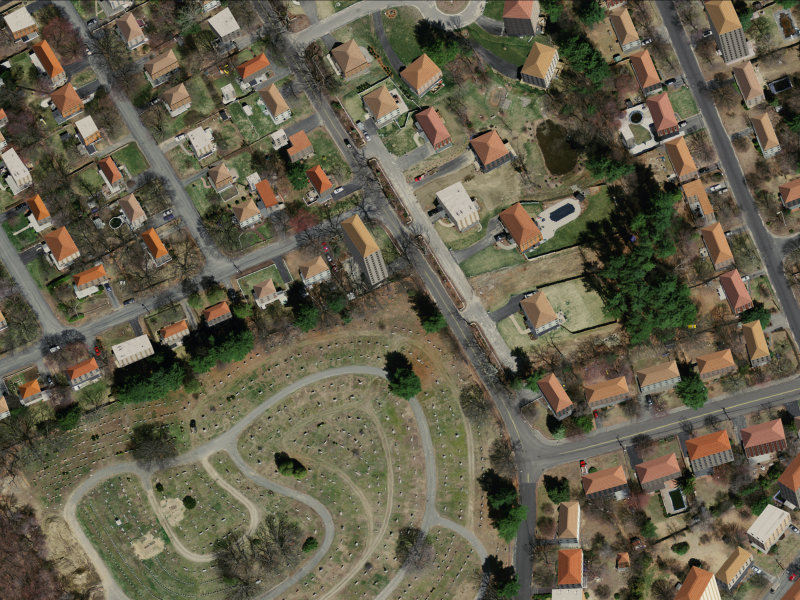}
%\begin{tabular}{cc}
%\includegraphics[scale=0.145]{data1.png}& %\includegraphics[scale=0.145]{data2.png}\\
%\vspace{.5mm}
%\end{tabular}
%\begin{tabular}{cc}
% \includegraphics[scale=.145]{data3.png}&
% \includegraphics[scale=.145]{data4.png}
%\end{tabular}
\caption{Some samples from our synthetic aerial dataset.}
\label{fig_graph8}
\end{figure}

\subsection{Dataset}
Whilst a limited number of datasets are available \cite{wamidatasets}, capturing Wide Area Motion Imagery can be an expensive and difficult process.  In addition to the obvious problems of obtaining specialist (often classified military) camera equipment, there is also the need to organize aircraft, pilots and permission to fly.  There are also broader legal implications of performing surveillance of a real city or community - a person could be identified by a house number he/she visits. Critically, there is no definitive form of ground truth (e.g., vehicle types \& positions) for comparative evaluation of methods - nor a means of placing a camera in a particular position, orientation and time, which is a useful capability for developing new algorithms.

A brief overview of the image and ground truth generation method is described here (see Figure 5), a more detailed discussion is presented in \cite{griffith2018}.

Firstly, the ground truth data is generated using a MATLAB simulation that controls an instance of SUMO traffic simulator (Simulation of Urban MObility) \cite{sumo}.  The MATLAB simulation can seamlessly transfer entities such as vehicles and people, between SUMO and itself.  SUMO is used to navigate vehicles as per a basic set of rules of the road, and also navigates pedestrians along sidewalks and crosswalks.  The MATLAB simulation handles the high level goals of the people such as when and where to visit (e.g., shop or workplace) the mode of transport to be used. It also manages other bespoke "`micro-simulations"' such as walking from the sidewalk to the building doorway, and people using bus stops. 

\begin{figure}[!t]
\centerline{\includegraphics[width=1.0\linewidth]{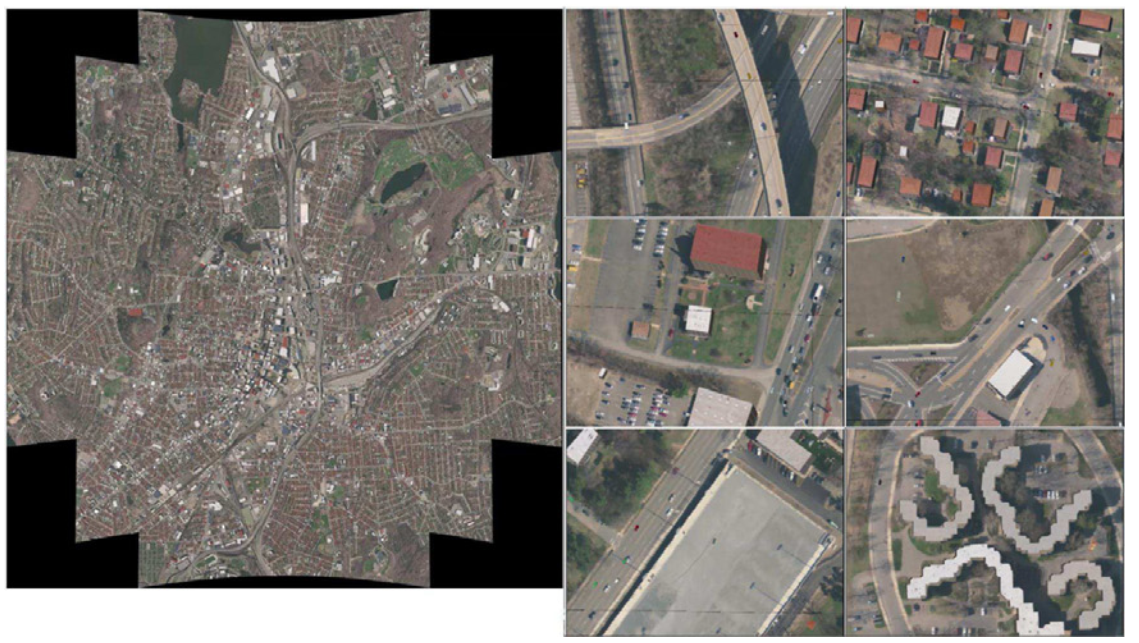}}
\caption{(left) ARGUS field of view illustrates the large area simulated (approx 6km x 6km). (right) 6 examples of  various areas at the same point in time.}
\label{fig}
\end{figure}

Similarly, images are then generated from the ground truth at each time step using a MATLAB controlled instance of the X-Plane flight simulator \cite{xplanewebsite}.  The MATLAB image generator can position the flight simulator's viewpoint, and spawn 3D vehicles of the correct type and color, finally triggering image captures and converting to the MATLAB matrix format (for saving or further processing).  A configuration based on the DARPA/BAE Systems ARGUS-IS imager \cite{argus} is used that contains an array of 368 5-Megapixel subcameras.  This generates an image of 1.8-Gigapixels (or 5-Gigabytes of uncompressed RGB imagery per frame) capturing a circular area of approximately 6km diameter (at a 6km altitude).  Equations describing the configuration of the subcamera array can be found in \cite{griffith2018}.  

The final output per frame (each timestep) is the ground truth data saved in an XML format, 368 PNG subcamera images each with an accompanying metadata file for the subcamera position and orientation.  The example applications within \cite{griffith2018} detail a tiled video playback tool, and an analysis tool for interpreting the ground truth data directly without the imagery (e.g., tracing paths taken by vehicles). \\

\subsection{Results}

In order to test our proposed method, we implemented two different experimental setups. First, we trained the proposed method on 75 percent of the dataset (all the 14 classes), which contains visual and textual information about the vehicle types and colors. Then, the remaining samples were used for the testing. Figure 6 shows the accuracy, true positive, and true negative percentages for the testing set. The three results indicate the promising power of the proposed method for recognizing the vehicles and their types and colors for the synthetic aerial images. 

To check the ability of the proposed method on unseen classes (open-ended setup), we repeated the experiment in a way that we only trained on 13 classes and set one class aside for testing. This experimental setup is repeated 14 times for all the 14 classes. Figure 7 shows the accuracy of the system for the unseen classes. Based on this experiment, we can see that the proposed method is capable of recognizing unseen vehicles but belonging to similar classes.

\begin{figure}[!t]
\centering
\includegraphics[scale=.25]{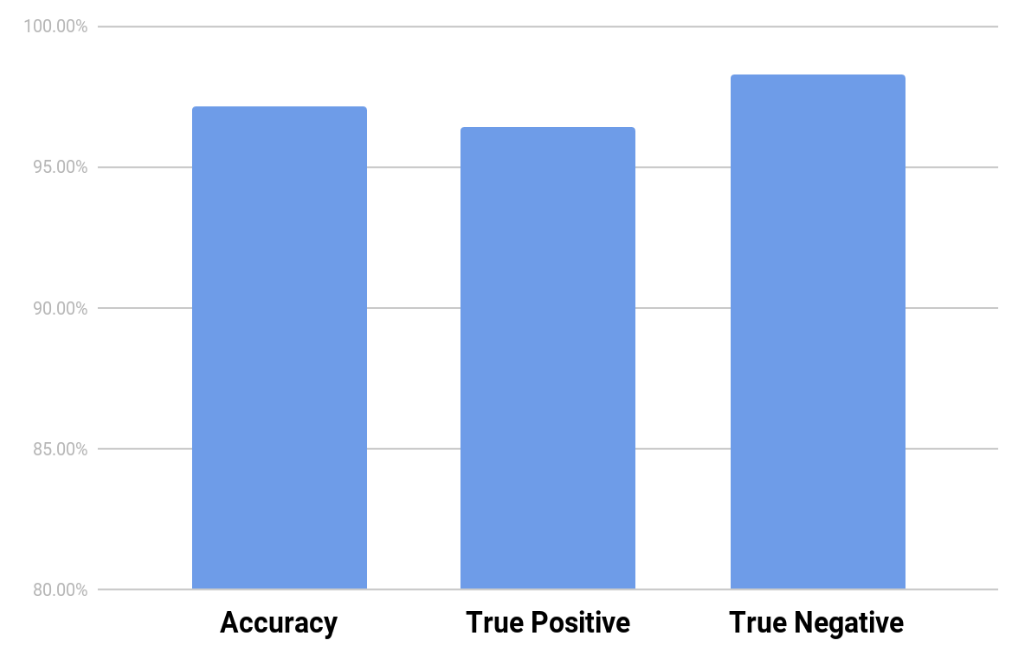}
\caption{The performance of the proposed method on the synthetic aerial dataset.}
\label{fig_graph4}
\end{figure}

In order to understand the underlying process in the system, we tried to visualize the textual information of the desired class labels in the hidden layer. Just for this experiment, we forced the fully connected layers for the textual feature to have a two dimensional representation. Figure 8 shows this two dimensional sub-space. It is clear that the two different vehicle types, trucks and cars, have a similar feature pattern and lie in a similar manifold. The colors of the vehicles in these two vehicle types are also in the same order. Therefore, it might be true that if the system is not trained on yellow cars, for example, it can respond to the unseen yellow car samples during the testing phase by considering the similar manifolds of the trucks and cars.\\

\section{Conclusion}
In this paper, we investigated the problem of aerial vehicle detection. We assumed that a deep detector with a fast and accurate performance like single shot multi-box detector or SSD generates a number of objects of interest for each aerial image, which can be called as image proposals or image patches. In the next step, we used a VGG-16 structure framework to extract visual information for the generated image proposals. On the other hand, the bag of words representation and fully-connected layers are used to make a textual feature representation for the desired classes.

The visual and textual information are fused and make a common latent sub-space, which is called visual-textual sub-space. Based on this sub-space a softmax classifier is trained to generate yes or no outputs that correspond to the cases when the input image patch has the desired class label or not. It is important to note that all the weights of the second step including the convolutional layers, the fully connected layers, and the softmax classifier, are optimized simultaneously. In other words, visual and textual features are trained together.

We tested our system on a synthetic aerial dataset that contains information about the vehicle types and vehicle colors. Results on this dataset, showed that in addition to the promising performance when recognizing seen or trained classes, our framework can recognize unseen classes as well, and this is the advantage of the open-ended framework. For the future works, collecting or synthesizing more complicated datasets that have both visual and rich textual information would result in the further improvements in the field. \\

\begin{figure}[!t]
\centering
\includegraphics[scale=.5]{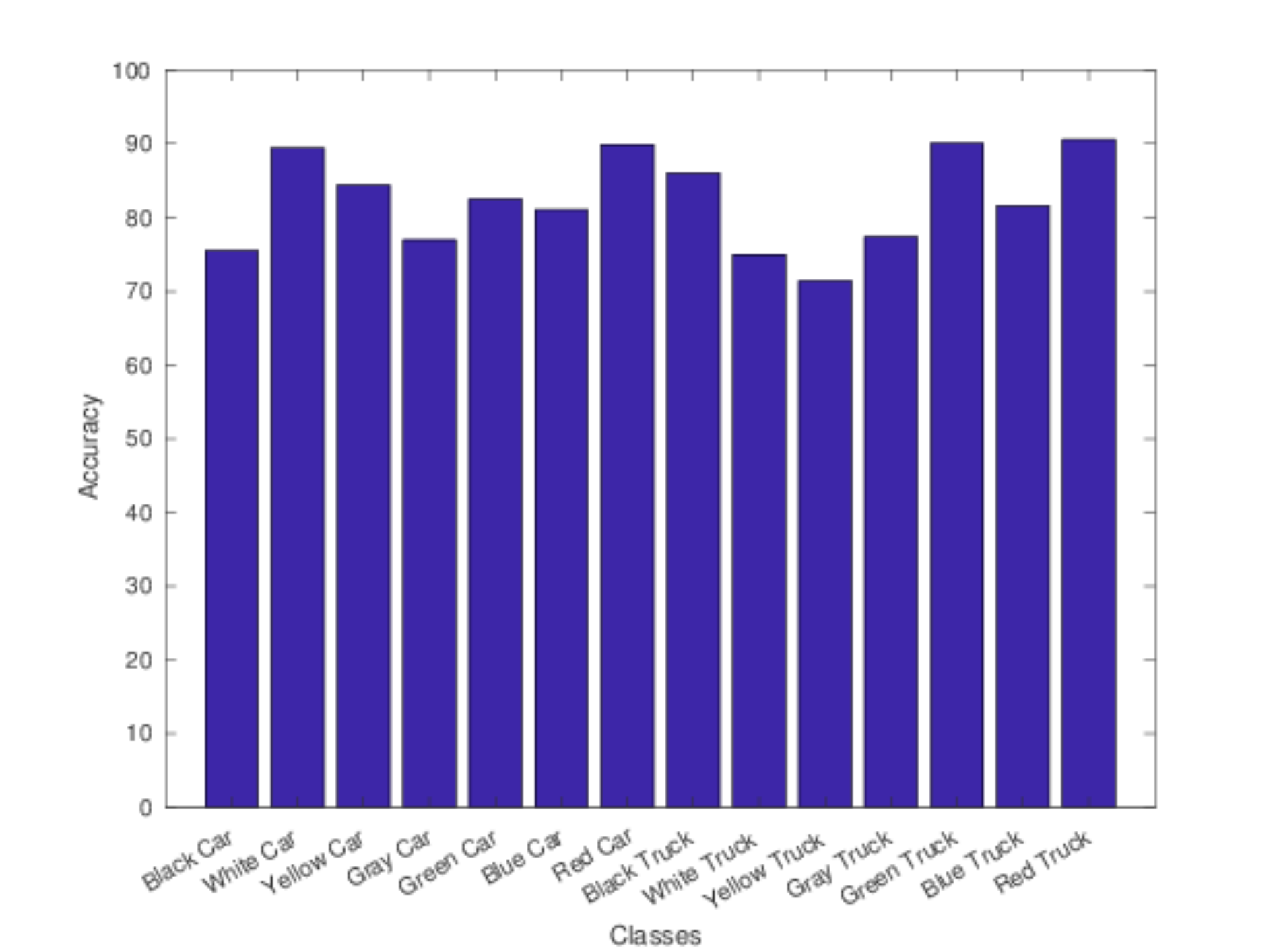}
\caption{The accuracy of the proposed method in the unseen experiment.}
\label{fig_graph5}
\end{figure}

\begin{figure}[t]
\centering
\includegraphics[scale=.25]{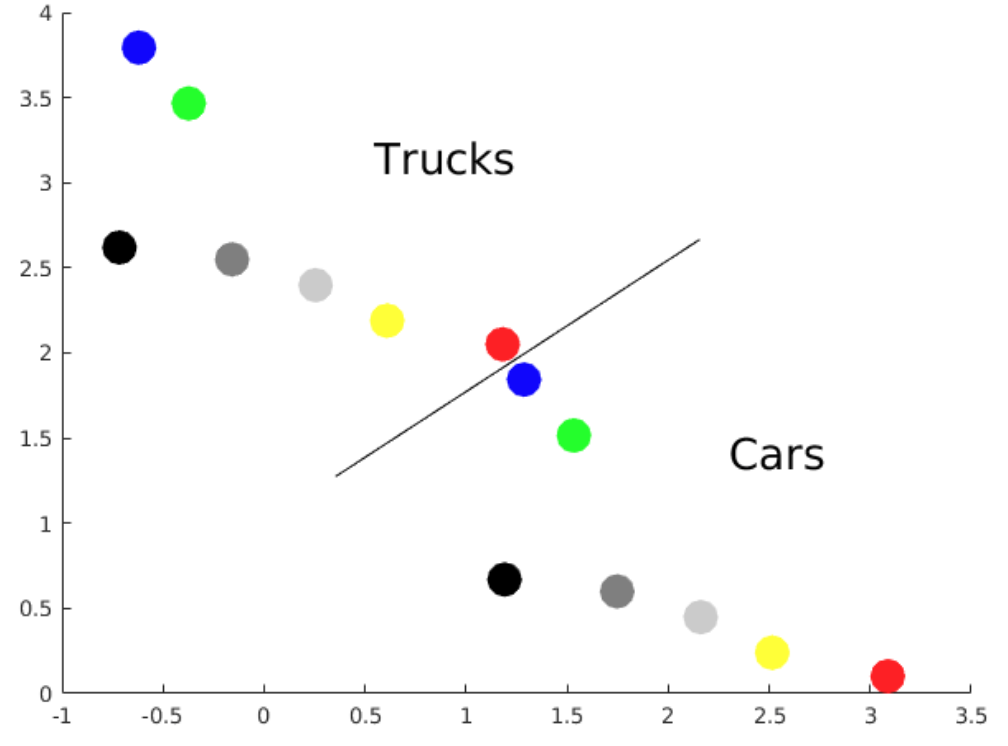}
\caption{2D visualization of the desired classes. The line separates the cars and trucks (vehicle types), and the colors correspond to the vehicle colors.}
\label{fig_graph9}
\end{figure}

\bibliography{Main.bib}

\begin{thebibliography}{10}
\providecommand{\url}[1]{#1}
\csname url@samestyle\endcsname
\providecommand{\newblock}{\relax}
\providecommand{\bibinfo}[2]{#2}
\providecommand{\BIBentrySTDinterwordspacing}{\spaceskip=0pt\relax}
\providecommand{\BIBentryALTinterwordstretchfactor}{4}
\providecommand{\BIBentryALTinterwordspacing}{\spaceskip=\fontdimen2\font plus
\BIBentryALTinterwordstretchfactor\fontdimen3\font minus
  \fontdimen4\font\relax}
\providecommand{\BIBforeignlanguage}[2]{{%
\expandafter\ifx\csname l@#1\endcsname\relax
\typeout{** WARNING: IEEEtran.bst: No hyphenation pattern has been}%
\typeout{** loaded for the language `#1'. Using the pattern for}%
\typeout{** the default language instead.}%
\else
\language=\csname l@#1\endcsname
\fi
#2}}
\providecommand{\BIBdecl}{\relax}
\BIBdecl

\bibitem{vqapaper}
S.~Antol, A.~Agrawal, J.~Lu, M.~Mitchell, D.~Batra, C.~Lawrence~Zitnick, and
  D.~Parikh, ``Vqa: Visual question answering,'' in \emph{Proceedings of the
  IEEE International Conference on Computer Vision}, 2015, pp. 2425--2433.

\bibitem{fusion}
A.~Soleimani and N.~Nasrabadi, ``Convolutional neural networks for aerial
  multi-label pedestrian detection,'' in \emph{21st International Conference on
  Information Fusion}, 2018.

\bibitem{ssd}
W.~Liu, D.~Anguelov, D.~Erhan, C.~Szegedy, S.~Reed, C.-Y. Fu, and A.~C. Berg,
  ``Ssd: Single shot multibox detector,'' in \emph{European conference on
  computer vision}.\hskip 1em plus 0.5em minus 0.4em\relax Springer, 2016, pp.
  21--37.

\bibitem{hog}
N.~Dalal and B.~Triggs, ``Histograms of oriented gradients for human
  detection,'' in \emph{Computer Vision and Pattern Recognition, 2005. CVPR
  2005. IEEE Computer Society Conference on}, vol.~1.\hskip 1em plus 0.5em
  minus 0.4em\relax IEEE, 2005, pp. 886--893.

\bibitem{sift}
D.~G. Lowe, ``Distinctive image features from scale-invariant keypoints,''
  \emph{International journal of computer vision}, vol.~60, no.~2, pp. 91--110,
  2004.

\bibitem{svm}
C.~Cortes and V.~Vapnik, ``Support-vector networks,'' \emph{Machine learning},
  vol.~20, no.~3, pp. 273--297, 1995.

\bibitem{voc}
M.~Everingham, L.~Van~Gool, C.~K. Williams, J.~Winn, and A.~Zisserman, ``The
  pascal visual object classes (voc) challenge,'' \emph{International journal
  of computer vision}, vol.~88, no.~2, pp. 303--338, 2010.

\bibitem{imagenet}
J.~Deng, W.~Dong, R.~Socher, L.-J. Li, K.~Li, and L.~Fei-Fei, ``Imagenet: A
  large-scale hierarchical image database,'' in \emph{Computer Vision and
  Pattern Recognition, 2009. CVPR 2009. IEEE Conference on}.\hskip 1em plus
  0.5em minus 0.4em\relax IEEE, 2009, pp. 248--255.

\bibitem{lenet}
Y.~LeCun, L.~Bottou, Y.~Bengio, and P.~Haffner, ``Gradient-based learning
  applied to document recognition,'' \emph{Proceedings of the IEEE}, vol.~86,
  no.~11, pp. 2278--2324, 1998.

\bibitem{alex}
A.~Krizhevsky, I.~Sutskever, and G.~E. Hinton, ``Imagenet classification with
  deep convolutional neural networks,'' in \emph{Advances in neural information
  processing systems}, 2012, pp. 1097--1105.

\bibitem{RCNN}
R.~Girshick, J.~Donahue, T.~Darrell, and J.~Malik, ``Rich feature hierarchies
  for accurate object detection and semantic segmentation,'' in
  \emph{Proceedings of the IEEE conference on computer vision and pattern
  recognition}, 2014, pp. 580--587.

\bibitem{selective}
J.~R. Uijlings, K.~E. Van De~Sande, T.~Gevers, and A.~W. Smeulders, ``Selective
  search for object recognition,'' \emph{International journal of computer
  vision}, vol. 104, no.~2, pp. 154--171, 2013.

\bibitem{fastrcnn}
R.~Girshick, ``Fast r-cnn,'' in \emph{Proceedings of the IEEE international
  conference on computer vision}, 2015, pp. 1440--1448.

\bibitem{faster}
S.~Ren, K.~He, R.~Girshick, and J.~Sun, ``Faster r-cnn: Towards real-time
  object detection with region proposal networks,'' in \emph{Advances in neural
  information processing systems}, 2015, pp. 91--99.

\bibitem{yolo}
J.~Redmon, S.~Divvala, R.~Girshick, and A.~Farhadi, ``You only look once:
  Unified, real-time object detection,'' in \emph{Proceedings of the IEEE
  conference on computer vision and pattern recognition}, 2016, pp. 779--788.

\bibitem{vgg}
K.~Simonyan and A.~Zisserman, ``Very deep convolutional networks for
  large-scale image recognition,'' \emph{arXiv preprint arXiv:1409.1556}, 2014.

\bibitem{okutama}
M.~Barekatain, M.~Mart{\'\i}, H.-F. Shih, S.~Murray, K.~Nakayama, Y.~Matsuo,
  and H.~Prendinger, ``Okutama-action: An aerial view video dataset for
  concurrent human action detection,'' in \emph{1st Joint BMTT-PETS Workshop on
  Tracking and Surveillance, CVPR}, 2017, pp. 1--8.

\bibitem{wamidatasets}
\BIBentryALTinterwordspacing
``{The Sensor Data Management System},'' 2016. [Online]. Available:
  \url{https://www.sdms.afrl.af.mil}
\BIBentrySTDinterwordspacing

\bibitem{griffith2018}
E.~J. Griffith, C.~Mishra, J.~F. Ralph, and S.~Maskell, ``A system for the
  generation of synthetic wide area aerial surveillance imagery,''
  \emph{Simulation Modelling Practice and Theory}, vol.~84, pp. 286--308, 2018.

\bibitem{sumo}
D.~Krajzewicz, E.~Brockfeld, J.~Mikat, J.~Ringel, C.~R{\"o}ssel,
  W.~Tuchscheerer, P.~Wagner, and R.~W{\"o}sler, ``Simulation of modern traffic
  lights control systems using the open source traffic simulation sumo,'' in
  \emph{Proceedings of the 3rd industrial simulation conference}, vol. 2205,
  2005.

\bibitem{xplanewebsite}
\BIBentryALTinterwordspacing
``{X Plane website},'' 2018. [Online]. Available: \url{https://www.x-plane.com}
\BIBentrySTDinterwordspacing

\bibitem{argus}
\BIBentryALTinterwordspacing
``{ARGUS-IS brochure-BAE systems},'' 2016. [Online]. Available:
  \url{http://www.baesystems.com/en/download-en/20151124113917/1434554721803.pdf}
\BIBentrySTDinterwordspacing

\end{thebibliography}
\bibliographystyle{IEEEtran}

\end{document}